\documentclass{article}

\usepackage{arxiv}

\usepackage[utf8]{inputenc} 
\usepackage[T1]{fontenc}    
\usepackage{url}            
\usepackage{booktabs}       
\usepackage{amsfonts}       
\usepackage{nicefrac}       
\usepackage{microtype}      
\usepackage{lipsum}		
\usepackage{graphicx}
\usepackage[square,numbers]{natbib}
\usepackage{doi}
\usepackage{amsmath}

\DeclareMathOperator*{\argmin}{arg\,min}
\usepackage{algorithm, algpseudocode}
\usepackage{svg}

\title{A new dynamical model for solving rotation averaging problem}

\author{  Zinaid Kapić, Aladin Crnkić \\
	Faculty of Technical Engineering \\
	University of Bihać \\
    Bihać \\
	\texttt{zinaid.kapic@unbi.ba, aladin.crnkic@unbi.ba} \\
	\And
	Vladimir Jaćimović, Nevena Mijajlović \\
    Faculty of Natural Sciences and Mathematics \\
    University of Montenegro \\
	Podgorica \\
	\texttt{vladimirj@ucg.ac.me, nevenami@ucg.ac.me} \\
}

\renewcommand{\shorttitle}{A new dynamical model for solving rotation averaging problem}

\hypersetup{
pdftitle={A new dynamical model for solving rotation averaging problem},
pdfsubject={Computer Science},
pdfauthor={Zinaid Kapic, Aladin Crnkic, Vladimir Jacimovic, Nevena Mijajlovic},
pdfkeywords={D rotations, rotation averaging, special orthogonal group, non-Abelian Kuramoto model},
}

\begin{document}
\maketitle

\begin{abstract}
	The paper analyzes the rotation averaging problem as a minimization problem for a potential function of the corresponding gradient system. This dynamical system is one generalization of the famous Kuramoto model on special orthogonal group SO(3), which is known as the non-Abelian Kuramoto model. We have proposed a novel method for finding weighted and unweighted rotation average. In order to verify the correctness of our algorithms, we have compared the simulation results with geometric and projected average using real and random data sets. In particular, we have discovered that our method gives approximately the same results as geometric average.
\end{abstract}


\section{Introduction}
Rotation averaging is probably one of the most important
problems in computer vision and robotics. This problem takes three different forms: single, multiple and conjugate rotation averaging. In the first problem, a single rotation is calculated from several measurements. In the second problem, absolute orientations are calculated from several relative orientation measurements, and the third problem relates to a pair of coordinate frames \cite{Hartley2013,Dellaert2020,Purkait2020}. 

The rotation averaging problem has gained significant
attention over the past few decades in many applications. Govindu in papers \cite{Govindu2001, Govindu2004, Govindu2006} introduces the application of rotation average for problems in the field of structure-from-motion. In paper \cite{Gramkow2001}, Gramkow compares three different methods for
calculating the average of rotations, relaying on rotation parameterization with unit quaternions, orthogonal rotational matrices and angle-axis representation. The most characteristic problem in this field is the integration of the position of the object measured with different cameras into a unique average position as in \cite{Dai2010, HongdongLi2008, Lebraly2010}. Humbert, Bingham and Bachmann apply rotational data to investigate the orientation of cubic crystals on the metal surface \cite{Humbert1996, Bachmann2010, Bingham2009}. Rotation averaging problem is also used in the study of geology \cite{Manning1996} or in genetics to model DNA \cite{Prentice1987}. 

The main problem with calculating the average of a set of rotations lies in the fact that rotations do not belong to a Euclidean space. Therefore, rotations represented in forms such as matrices, Euler angles, and unit quaternions are constrained to some nonlinear manifolds. Each of these representations has its preferred application. Euler angles are commonly used in robotics \cite{craig1986introduction}, while unit quaternions and matrices have applications in computer vision and computer graphics \cite{Pervin1982, Shoemake1985, Heeger}. We have used 3x3 real orthogonal matrices with determinant 1, known as rotation matrices. The set of all such matrices form a special orthogonal group SO(3). 

The paper examines so called single rotation averaging problem. The goal is to find an average (mean) of $N$ rotation matrices $R_1,\dots,R_N \in SO(3)$  that minimizes the sum of squared distances from that rotation matrix to the given rotation matrices $R_1,\dots,R_N$, as follows \cite{Moakher2002}:
\begin{equation}
M(R_1,\dots,R_N):= 
\argmin_{R_1\subset SO(3)}
\sum_{n=1}^{N} {d(R_n,R)^{2}}
\end{equation}
Here $d(\cdot, \cdot)$ represents a metric (distance) on $SO(3)$. Subsequently, we will define two rotation averages with two different choices of metric functions. We will start with the projective arithmetic average which is associated with the following metric \cite{Moakher2002}:
\begin{equation}
    d_F(R_1, R_2)=||R_1-R_2||_F,
\end{equation}
where $||\cdot||_F$ represents Frobenius norm. Another interesting average with respect to metric, defined by 
\begin{equation}
    d_R(R_1, R_2)=\frac{1}{\sqrt{2}}||Log({R}_{1}^{T}R_2||_F,
\end{equation}
is geometric average. The metric (2) is called geodesic because it represents the length of the shortest geodesic curve that connects rotations $R_1$ and $R_2$ \cite{Moakher2002}.

We want to find weighted rotation average in many applications. Some of those applications are spacecraft attitude estimation \cite{Markley2007} or genetics for DNA modelling \cite{Manning1996}. The average of N given rotations R1,…,RN with weights $\kappa = (\kappa_1,\dots,\kappa_N)$, according to \cite{Moakher2002}, is defined as:
\begin{equation}
M_W(R_1,\dots,R_N;\kappa):= 
\argmin_{R_1\subset SO(3)}
\sum_{n=1}^{N} {\kappa_n d(R_n,R)^{2}}.
\end{equation}
Here we also use metrics (1) and (2) to calculate projective arithmetic and geometric weighted average.

The paper proposes a new method for finding rotation average based on generalizations of the Kuramoto model to higher dimensions. Kuramoto model \cite{Kuramoto} is the most significant model in studying the collective behavior and self-organization in large populations of coupled oscillators. The first generalization of this model, known as the non-Abelian Kuramoto model, has been introduced by Lohe \cite{Lohe2009, Jaimovi2018}:
\begin{equation}
i\dot{U_j}U_j^*=H_j-\frac{iK}{2N}\sum_{i=1}^{N} {(U_jU_i^*-U_iU_j^*), j=1, \dots, N.}
\end{equation}
This model is an extension of classical Kuramoto model on the group of the unitary matrix U(n). 

Different variations of the Kuramoto model and its generalizations have found application in many scientific disciplines in solving various problems, such as community detection in complex networks \cite{Arenas2008, Arenas2006}, clustering of static and stream data \cite{Crnki2018, Crnki2020}, coordination in multi-agent systems \cite{Crnki2020SO3}, and other applications in science and engineering \cite{DazGuilera2008}. 

In the next section, we will introduce a model for solving unweighted and weighted rotation average problems and explain our algorithms in detail. In Section 3 we will present simulation results that illustrate our method on real and random data sets. Finally, in Section 4, we will briefly discuss the results, draw conclusions, and point out possible disadvantages and potential upgrades of our methods. 
\section{Algorithm}
In this section we will formulate rotation average problem as optimization problem of potential function
\begin{equation}
P(R_1(t), \dots, R_N(t))=-\frac{1}{2N^2} \sum_{i=1}^{N} {\sum_{j=1}^{N}{Tr(R_i^*(t)R_j(t))} \to MIN }
\end{equation}
with respecct to $R_j(t), j=1, \dots, N$. The notation $R^*$ stands for the transpose of a matrix $R$. The following dynamical system
\begin{equation}
\frac{d}{dt}R_j(t)=\frac{1}{N}\sum_{i=1}^{N}{(R_i(t)-R_j(t)R_i^*(t)R_j(t))}, j=1, \dots N,
\end{equation}
is obtained as gradient descent method for minimization problem (3). Notice that (4) preserves $SO(3)$, i.e. if the initial condition satisfy $R_j(0) \in SO(3)$, it stands that $R_j(t) \in SO(3)$ for all t. It is obvious that alignment (synchronization) $R_1=R_2= \dots =R_N$ of the population is global minimum for potential function P. In this layout, this equilibrium configuration corresponds to average of rotations. 

Similarly, we can easily introduce weighted rotation average problem as minimization of function
\begin{equation}
P_W(R_1(t), \dots, R_N(t))=-\frac{1}{2N^2} \sum_{i=1}^{N} {\sum_{j=1}^{N}{\kappa_i Tr(R_i^*(t)R_j(t))} \to MIN }
\end{equation}
Here $\kappa_i, i=1,2, \dots ,N$ represents weights of rotations.

The corresponding gradient system reads as: 
\begin{equation}
\frac{d}{dt}R_j(t)=\frac{1}{N}\sum_{i=1}^{N}{\kappa_i(R_i(t)-R_j(t)R_i^*(t)R_j(t))}, j=1, \dots N,
\end{equation}

Now we will explain our method for finding rotation average through algorithms in more details.

First, we will introduce algorithm for solving unweighted rotation average. We refer to this algorithm as KL algorithm (after Kuramoto and Lohe). Let us primarily suppose that the data set contains $N$ rotations $R_i', i=1,2, \dots,N$ represented by $SO(3)$ matrices. The KL algorithm is as follows:

\begin{algorithm}
\caption{KL algorithm for computing unweighted rotation average}
\begin{algorithmic}[1]
\State Enter $N$, $R_{i}^{'}$
\State Choose tolerance $\varepsilon$, step $\delta$, and define $T=0$
\State Solve (7) with $R_i(0)=R_i'$
\State Calculate $\hat{R}(t)=\frac{1}{N}\sum_{i=1}^{N}{R_i(t)}$, for $t>0$
\Loop
  \If{$1-det\hat{R}(T+\delta)<\varepsilon$}
    \State \textbf{return} $R(R_1', \dots, R_2')=R_i(T)$, for any $i$
  \Else
    \State $T=T+\delta$
  \EndIf
\EndLoop
\end{algorithmic}
\end{algorithm}

In addition to KL algorithm, we have also introduced KLW algorithm for solving weighted rotation average. The complete algorithm is as follows: 
\begin{algorithm}
\caption{KLW algorithm for computing weighted rotation average}
\begin{algorithmic}[1]
\State Enter $N$, $R_{i}^{'}, \kappa_i$
\State Choose tolerance $\varepsilon$, step $\delta$, and define $T=0$
\State Solve (9) with $R_i(0)=R_i'$ and $\kappa_i$
\State Calculate $\hat{R}(t)=\frac{1}{N}\sum_{i=1}^{N}{R_i(t)}$, for $t>0$
\Loop
  \If{$1-det\hat{R}(T+\delta)<\varepsilon$}
    \State \textbf{return} $R_W(R_1', \dots, R_2'; \kappa)=R_i(T)$, for any $i$
  \Else
    \State $T=T+\delta$
  \EndIf
\EndLoop
\end{algorithmic}
\end{algorithm}
It is noticeable that step 4 of both algorithms defines the Euclidean average $\hat{R}$. This matrix does not belong to special orthogonal group $SO(3)$. However, determinant of matrix $\hat{R}$ represents the global order parameter (a measure of alignment) of the population, and its value is between 0 and 1, i.e. $0\leq det\hat{R} \leq 1$. In case when $det\hat{R}=1$, we get the full alignment of the population, which corresponds to the rotation average. 
\section{Simulations}
This section presents simulation results of our method on one real and two randomly generated data sets. The average obtained with our method on these data sets was verified by comparison with the projective and geometric average. 

All simulations of our algorithms and well-known algorithms used for comparison are implemented using the Wolfram Mathematica package. According to KL and KLW algorithms, we have entered a number $N$ of rotations and expressed those rotations in $SO(3)$ form. Each simulation uses tolerance $\varepsilon = 10^{-5}$ and step $\delta=0.01$. Using inputs from algorithms step 1 and step 2, the program solves dynamical systems (7) and (9) for KL and KLW algorithm, respectively. To solve these systems, we have used the 4th order Runge-Kutta method. The Euclidean average is calculated in step 4, which is required in the next step for the stopping criteria. When the condition from step 5 is true (alignment occurred), algorithms return a value of rotation or weighted rotation average. However, it can return a new step for each iteration until the alignment of the population occurs.

As an example, we can use the fact that rotation matrices are orthogonal and every column of the rotation matrix has a length equal to 1. That means that every rotation can be represented as a point on the surface of the unit sphere $S^2$. Mathematically expressed, if $X\in SO(3)$ then $X_{V_1},X_{V_2}, X_{V_3} \in S^2$ where $V_1=[1,0,0]^T, V_2=[0,1,0]^T, V_3=[0,0,1]^T$ are unit vectors.

As the first example, we have analyzed the real data set Drill. This data assesses variations of human movements while performing a task. It is collected by monitoring eight subjects drilling into a metal plate using infrared cameras. The data is used to represent the orientation of each subject’s wrist, elbow, and shoulder in one of six positions \cite{Rancourt2000}. For this simulation, we have used 60 rows of drill data that belong to Subject “1” and is performed by the subject’s elbow.

Using the KL algorithm, we have obtained the average of this data in the moment $T=5.41$. Fig 1. illustrates the evolution of the potential function $P(R_1(t),\dots, R_N(t))$ for $t\in[0,8]$.

\begin{figure}[!htbp]
	\centering
    \includegraphics[width=0.65\columnwidth]{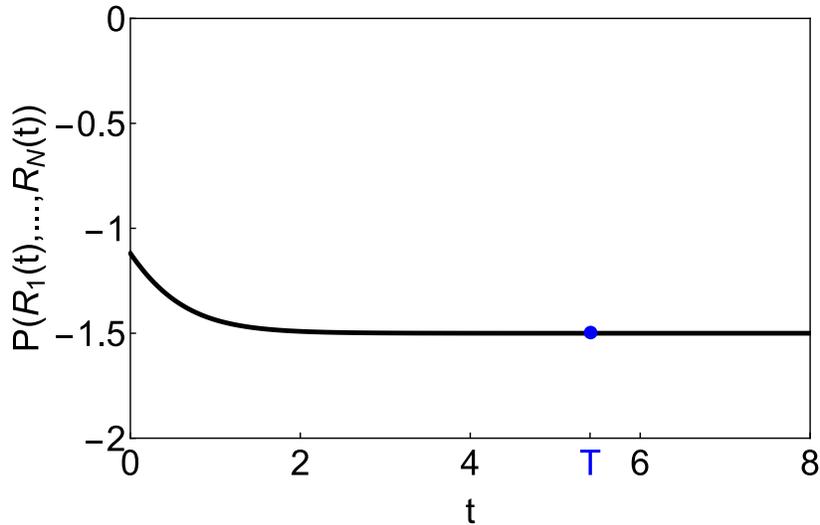}
	\caption{The evolution of the potential function (6) for the real data set Drill.}
	\label{fig:fig1}
\end{figure}

The data set of rotations and its averages are shown in Fig 2b), as points on the unit sphere. With closer examination, it can be seen that geometric and KL rotation average are almost identical and overlap, while the projected average deviates slightly from them. That, of course, confirmed the average values provided in Table 1. Fig. 2a) illustrates the evolution of the order parameter $det\hat{R}(t)$.

\begin{table}[!htbp]
	\caption{Rotation averages obtained for data set Drill using different algorithms}
	\centering
	\begin{tabular}{cc}
		\toprule
		\cmidrule(r){1-2}
		\textbf{Rotation average}     & \textbf{Result} \\
		\midrule
		\\
		Projected average & $\begin{bmatrix}
                            0.948745 & 0.307382 & 0.0734808 \\
                            0.229426 & -0.509944 & -0.829048 \\
                            -0.217364 & 0.803414 & -0.554328
                            \end{bmatrix}$       \\ \\
		Geometric average & $\begin{bmatrix}
                            0.947201 & 0.311427 & 0.0763086\\
                            0.227146 & -0.48376 & -0.845211\\   
                            -0.226306 & 0.817918 & -0.528957
                            \end{bmatrix}$      \\ \\
		KL average      &   $\begin{bmatrix}
                            0.947206 & 0.311415 & 0.0762942\\
                            0.227135 & -0.483792 & -0.845196\\
                            -0.226296 & 0.817904 & -0.528984
                            \end{bmatrix}$         \\ \\
		\bottomrule
	\end{tabular}
	\label{tab:table1}
\end{table}
The second data is sampled from the von Mises-Fisher distribution on $S^3$ with the mean direction $\mu=(1/2,1/2,1/2,1/2)$ and fixed concentration parameter $\kappa=0.5$. We converted this data to rotation matrices via the double cover map $S^3 \to SO(3)$. Collected data consists of 500 rotation matrices. KL average is obtained in the moment $T=6.48$. Simulation results on this data set are presented in Fig. 3 and Table 2.

\begin{table}[!htbp]
	\caption{Rotation averages obtained for random data sets using different algorithms}
	\centering
	\begin{tabular}{cc}
		\toprule
		\cmidrule(r){1-2}
		\textbf{Rotation average}     & \textbf{Result} \\
		\midrule
		\\
		Projected average & $\begin{bmatrix}
                            0.995009 & 0.00910849 & -0.0993697 \\
                            -0.00622441 & 0.999551 & 0.0292952 \\
                            0.099592 & -0.0285305 & 0.994619
                            \end{bmatrix}$       \\ \\
		Geometric average & $\begin{bmatrix}
                            0.999227 & 0.0208533 & -0.0333122 \\
                            -0.0203225 & 0.999662 & 0.0161936 \\
                            0.0336386 & -0.0155041 & 0.999314
                            \end{bmatrix}$      \\ \\
		KL average      &   $\begin{bmatrix}
                            0.999264 & 0.0201789 & -0.0326295 \\
                            -0.019653 & 0.999673 & 0.0163576 \\
                            0.0329489 & -0.0157043 & 0.999334
                            \end{bmatrix}$        \\ \\
		\bottomrule
	\end{tabular}
	\label{tab:table2}
\end{table}

KLW algorithm for calculating weighted rotation average is  verified with the previous algorithms on random data set. The data set consists of 300 rotation matrices, sampled in the same way as the second data set, with corresponding weight values. The weights are generated randomly in the range $[0,1]$. Using this data set, we have obtained the KLW average in the moment $T=13.38$. Simulation results are illustrated in Fig. 4 and Table 3.

\begin{table}[!htbp]
	\caption{Weighted rotation averages obtained for random data sets using different algorithms}
	\centering
	\begin{tabular}{cc}
		\toprule
		\cmidrule(r){1-2}
		\textbf{Rotation average}     & \textbf{Result} \\
		\midrule
		\\
		Projected average & $\begin{bmatrix}
                            0.99788 & 0.0573859 & 0.0307056 \\
                            -0.058145 & 0.998009 & 0.024426 \\
                            -0.0292428 & -0.0261596 & 0.99923
                            \end{bmatrix}$       \\ \\
		Geometric average & $\begin{bmatrix}
                            0.987262 & 0.0879129 & 0.13261 \\
                            -0.0897579 & 0.995932 & 0.00798791 \\
                            -0.131368 & -0.0197889 & 0.991136
                            \end{bmatrix}$      \\ \\
		KL average      &   $\begin{bmatrix}
                            0.989628 & 0.0750597 & 0.122484 \\
                            -0.0768147 & 0.996999 & 0.0096628 \\
                            -0.121391 & -0.0189712 & 0.992423 
                            \end{bmatrix}$         \\ \\
		\bottomrule
	\end{tabular}
	\label{tab:table3}
\end{table}

\begin{figure}[!htbp]
	\centering
    \includegraphics[width=1.0\columnwidth]{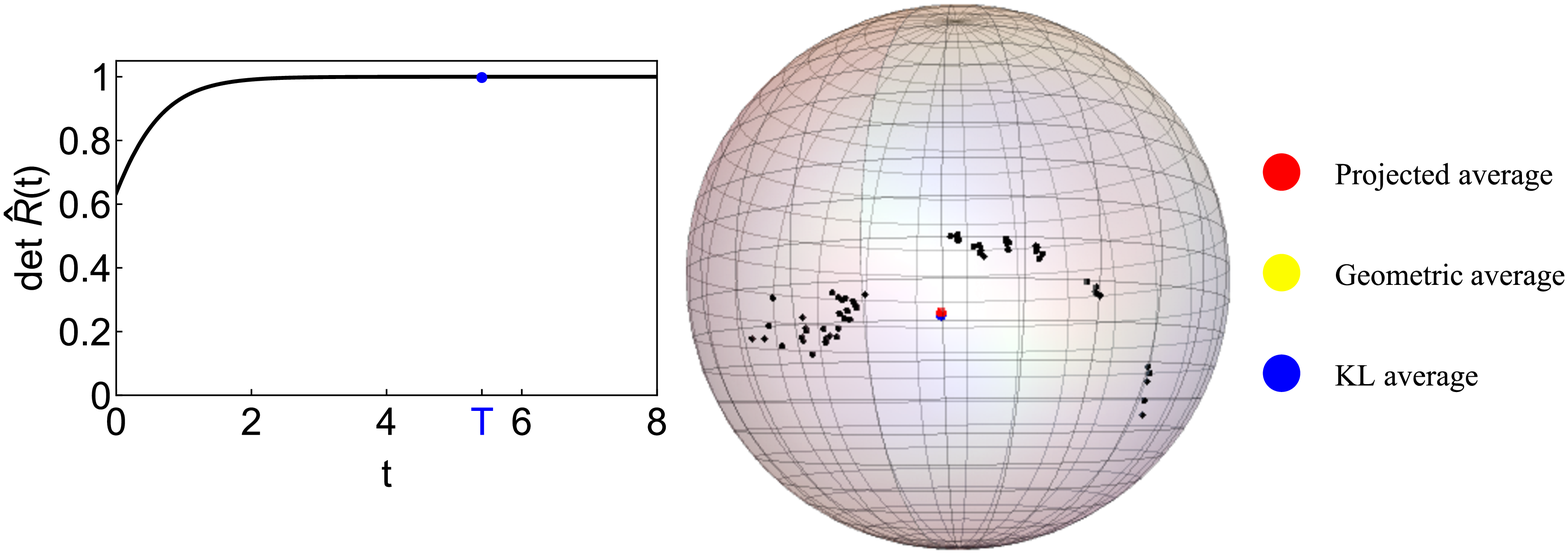}
	\caption{ The evolution of the order parameter $det\hat{R}(t)$ and visualization of rotations and its averages on unit spheres by plotting points on the unit sphere for the real data set Drill.}
	\label{fig:fig2}
\end{figure}

\begin{figure}[!htbp]
	\centering
    \includegraphics[width=1.0\columnwidth]{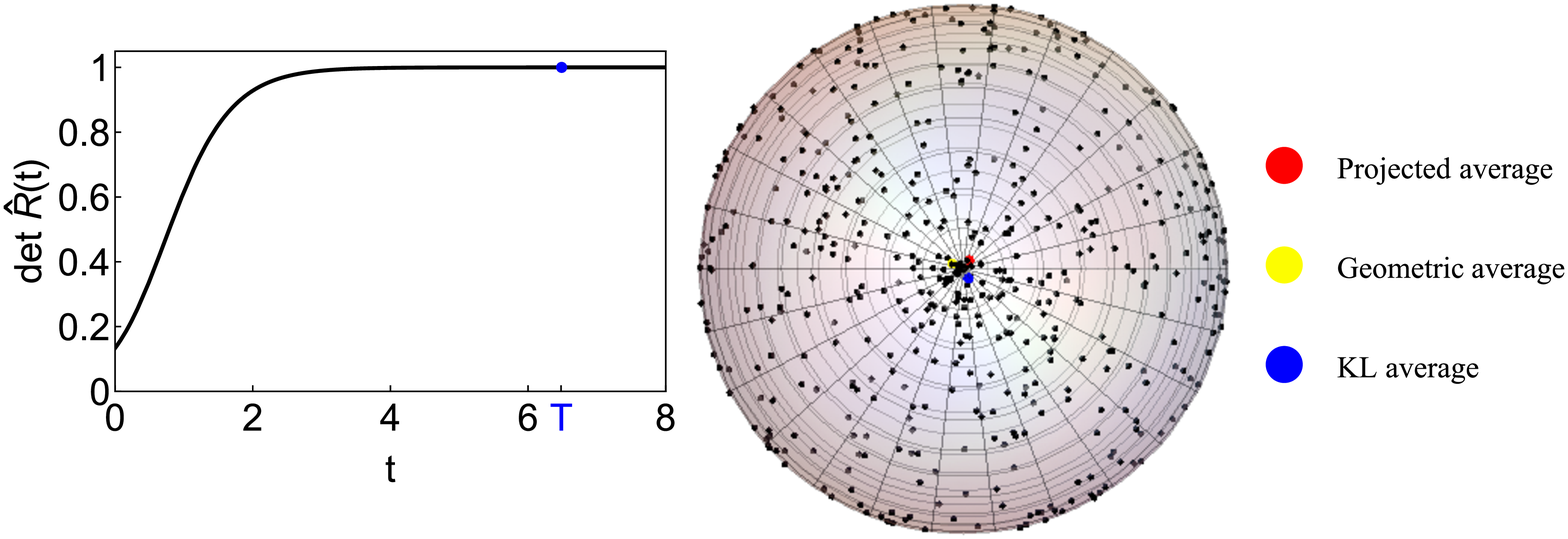}
	\caption{The evolution of the order parameter $det\hat{R}(t)$ and visualization of rotations and its averages on unit spheres by plotting points on the unit sphere for the random data set. }
	\label{fig:fig3}
\end{figure}

\begin{figure}[!htbp]
	\centering
    \includegraphics[width=1.0\columnwidth]{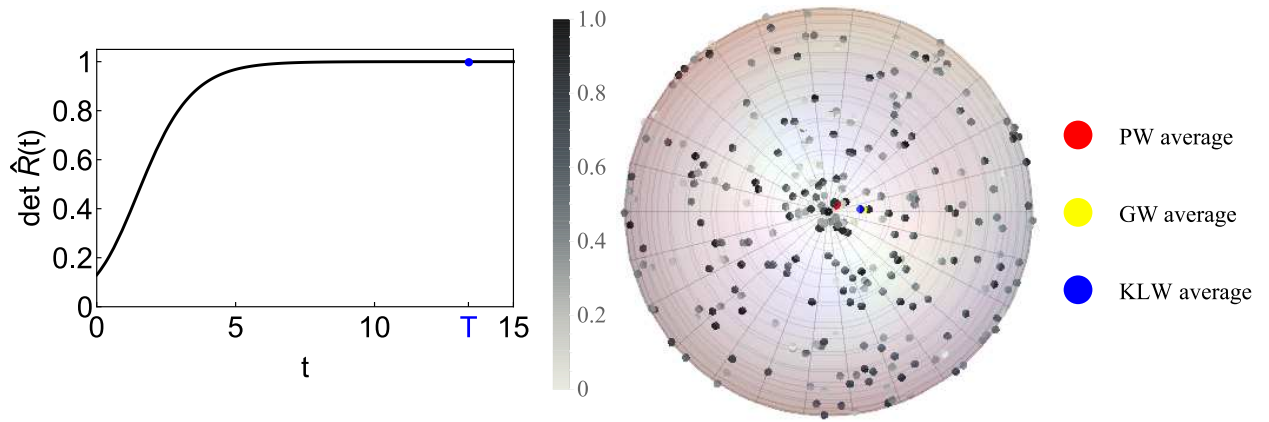}
	\caption{The evolution of the order parameter $det\hat{R}(t)$ and visualization of weighted rotations and its averages on unit spheres by plotting points on the unit sphere for the random data set. }
	\label{fig:fig4}
\end{figure}

\section{Conclusion}
The paper suggests the new method for calculating rotation average and weighted rotation average. This method is based on so-called non-Abelian Kuramoto model. The paper explains our algorithm and verifies it with the previous algorithms on real and random data sets. The simulation results indicate that our algorithm gives approximately the same results as geometric average. 

Proposed algorithms in this paper can be used, for instance, in a calibrated network of multiple cameras, where we can reduce the noise of single measurements by finding an average of the measured object orientations. 

The main disadvantage of this model is its inapplicability to huge data sets, because solving a large system of matrix ordinary differential equations represents a degradation in terms of speed and storage. 

It would be interesting to implement these algorithms in ROS (Robot Operating System) in the future research activities. ROS is a meta operating flexible system for writing robot software that offers users a way to quickly build, maintain and expand their robot, allowing them hardware abstraction for robot programming. For more details see \cite{Estefo2019}. 

Likewise, it is worth mentioning that the dynamical systems we introduced here allow for various modifications.Specifically, they can be easily adapted for interpolation 
between two rotations. We will address this possibility in further research.

\bibliographystyle{unsrtnat}







\end{document}